\documentclass[conference]{IEEEtran}
\IEEEoverridecommandlockouts

\usepackage{cite}
\usepackage{hyperref}
\usepackage{microtype}
\usepackage{booktabs}
\usepackage{amsmath,amssymb,amsfonts}
\usepackage{algorithmic}
\usepackage{graphicx}
\usepackage{textcomp}
\usepackage{xcolor,balance}
\def\BibTeX{{\rm B\kern-.05em{\sc i\kern-.025em b}\kern-.08em
    T\kern-.1667em\lower.7ex\hbox{E}\kern-.125emX}}

\begin{document}

\title{Forecasting with Multiple Seasonality}

\author{\IEEEauthorblockN{Tianyang Xie}
\IEEEauthorblockA{\textit{School of Statistics} \\
\textit{University of Minnesota Twin Cities}\\
Minneapolis, US \\
tianyang.xie.xty@gmail.com}
\and
\IEEEauthorblockN{Jie Ding}
\IEEEauthorblockA{\textit{School of Statistics} \\
\textit{University of Minnesota Twin Cities}\\
Minneapolis, US \\
dingj@umn.edu}}

\maketitle

\begin{abstract}
An emerging number of modern applications involve forecasting time series data that exhibit both short-time dynamics and long-time seasonality. Specifically, time series with multiple seasonality is a difficult task with comparatively fewer discussions. In this paper, we propose a two-stage method for time series with multiple seasonality, which does not require pre-determined seasonality periods. In the first stage, we generalize the classical seasonal autoregressive moving average (ARMA) model in multiple seasonality regime. In the second stage, we utilize an appropriate criterion for lag order selection. Simulation and empirical studies show the excellent predictive performance of our method, especially compared to a recently popular `Facebook Prophet' model for time series.
\end{abstract}


\section{Introduction}
In time series, seasonality is defined as the presence of variations that occur at specific regular intervals. Forecasting on time series with multiple seasonality that has different lengths of seasonality cycles is usually considered a difficult task. Therefore, detection and accommodation of the seasonality effect play an important role in time series forecasting.

Among all the forecasting techniques, the seasonal ARIMA model~\cite{SARIMA} and exponential smoothing technique ~\cite{winter, HOLT} are the classical approaches. The basic forms of these methods are only suited in modeling single seasonality, and unable to account for multiple seasonal patterns. Many studies have been conducted to extend the classical statistical forecasting models to accommodate multiple seasonal patterns. Representative methods include the double seasonal ARIMA model and exponential smoothing technique adapted from the simple Holt-Winters method~\cite{doubleSARIMA}, and the hidden Markov model with multiple seasonality~\cite{GOULD2008207}. However, existing techniques tend to be sensitive to over-parameterization and optimization issues, and are also unable to model complex seasonal patterns in a time series, as pointed out in~\cite{tbats}. A more flexible method from the state space perspective was developed in~\cite{tbats}, which is the state-of-the-art method for multiple seasonality. 

Recently, a new model named Facebook Prophet~\cite{prophet} is introduced that utilizes the generalized additive model~\cite{GAM}, and succeeds in forecasting Facebook's business data. In parallel to this development, neural networks (NNs) have been advocated as a strong alternative for forecasting time series with multiple seasonality. The recurrent neural network (RNN)~\cite{RNN} based deep learning techniques such as long short-term memory (LSTM)~\cite{LSTM}, gated recurrent unit (GRU)~\cite{GRU} achieve promising results in many real data applications~\cite{deep1,deep2}. 
In addition to the development in neural networks, hybridization of classical methods and deep learning has also been advocated. For example, in the winning submission of the M4 forecasting competition~\cite{MAKRIDAKIS2018802}, a hybrid approach called exponential smoothing-recurrent neural network (ES-RNN) succeeds in forecasting hourly time series data. 

Most of the classical parametric methods require the pre-specification of seasonality periods. For example, we usually fix the seasonality period to be 24 for hourly data, then we train model or employ seasonal adjustment on the belief of that. However, in multiple seasonality situation, the pre-specification can be risky when the data doesn't exhibit clear seasonal patterns. An example concerning the total sunspot number is included in Section~\ref{sub_sunspot}. Unfortunately, for methods trained in state-space innovative algorithms, model selection for seasonality periods is computationally prohibited. 
On the other hand, though deep learning techniques are free from pre-specification, they require a sophisticated tuning process and large datasets. 
Moreover, it is not clear how to extract the information of seasonality from existing deep learning architectures. 

This paper introduces a simple yet powerful method named Multiple Seasonality (MS) modeling procedure. The proposed method is to forecast centered or detrended time series with multiple seasonality by two stages. In the first stage, it detects seasonal components and fits generalized seasonal ARMA models. In the second stage, it selects an appropriate model and provides forecasting results. Instead of requiring the pre-specification of seasonality periods, it only needs to know how many seasonality components to include in the model. The procedure has a reasonable number of hyper-parameters, enabling easy interpretation and tuning compared with existing approaches. Simulation and empirical study show the superior performance of our method over the Facebook Prophet model and TBATS (which requires pre-specified seasonality periods). We have developed a comprehensive R package with documentation and plan to publish on CRAN.

The paper is outlined below. In Section II, we generalize the classical seasonal ARMA model and propose the MS modeling procedure followed by discussions. In Section III, we compare MS modeling procedure with other methods by various simulations and real-data case studies and show its suitability in modeling multiple seasonality.

\section{Method}

\subsection{Metric}

Observing a time series $X_t,t=1,\dots,N$ where N is the sample size, our goal is to perform n-step ahead prediction $\hat X_{N+n}$ with minimized cumulative mean squared error (CMSE).

\begin{equation}
\widehat{CMSE}_n = \frac{1}{n} \sum_{j=1}^n (X_{N+j} - \hat X_{N+j})^2 \notag
\end{equation}

\subsection{Model Formulation}

We first generalize the classical multiplicative seasonal ARMA model~\cite{SARIMA} with normal distributed noise. The model has pre-determined constants: $r$, the number of seasonality components; $\tau$, the maximum neighborhood length of each seasonality component; $\bar p$, the maximum AR components allowed; $\bar q$, the maximum MA components allowed. It also has parameters $p,q,C$ which need to be selected by an information criterion. The model is formed as following:
\begin{align}
X_t = \Phi' X_{t-1}^{t-p} + &\Psi' \epsilon_{t-1}^{t-q} + \sum_{i=1}^{r} (\Gamma_{i}' X_{\textbf{S}_i} + \Lambda_{i}' \epsilon_{\textbf{S}_i}) + \epsilon_t 
\end{align}

The observed data is $X_t \in \mathbb{R}, \text{ where }t=1,\dots,N$, and auto-regressive component is defined as $X_{t-1}^{t-p} = (X_{t-1},\dots,X_{t-p})'$, with coefficients $\Phi = (\phi_1,\dots,\phi_p)' \in \mathbb{R}^p$; The noise random variable is assumed as normal distributed $\epsilon_t {\sim} ^{i.i.d} N(0,\sigma^2) \in \mathbb{R}$, and moving average component is defined as $\epsilon_{t-1}^{t-q}=(\epsilon_{t-1},\dots,\epsilon_{t-q})' $, with coefficients $\Psi = (\psi_1,\dots,\psi_q)' \in \mathbb{R}^q$; Seasonality periods' collection $C$ is a collection of sets $C=\{ \textbf{S}_i, \text{ for }1\leq i \leq r\}$, where $r=|C|$ (the number of seasonality components) is pre-determined. $\textbf{S}_i$ is defined as set of neighboring integers where $\textbf{S}_i = \{t-S_{i,1},\dots,t-S_{i,\tau}\}$, $\tau=|\textbf{S}_i|$ (the neighborhood length of each seasonality component) is pre-determined for all $1\leq i \leq r$. Accordingly, we let 
\begin{align*}
&X_{\textbf{S}_i} = (X_{t-S_{i,1}},\dots,X_{t-S_{i,\tau}})' \in \mathbb{R}^{\tau} \\
&\Gamma_{i} = (\gamma_{i,1},\dots,\gamma_{i,\tau})' \in \mathbb{R}^{\tau}\\
&\epsilon_{\textbf{S}_i}=(\epsilon_{t-S_{i,1}},\dots,\epsilon_{t-S_{i,\tau}})' \in \mathbb{R}^{\tau} \\
&\Lambda_{i} = (\lambda_{i,1},\dots,\lambda_{i,\tau})' \in \mathbb{R}^{\tau}
\end{align*}
To exemplify the formulation, we take the selected model in Facebook Events Data case study in Section~\ref{subsec_face} as an example. Three seasonality components are assumed in the model ($r=3$), since more than two seasonality are suspected in the original paper that include the data~\cite{prophet}. The neighborhood length is allowed to be 13 ($\tau=13$). Parameters are selected as $p=3,q=4,C=\{(116,\dots,128),(278,\dots,290),(360,\dots,372)\}$. The model is written as:
\begin{align}
X_t &= \Phi' X_{t-1}^{t-3} + \Psi' \epsilon_{t-1}^{t-4} + (\Gamma_{1}' X_{t-116}^{t-128} + \Lambda_{1}' \epsilon_{t-116}^{t-128}) \notag \\
& \quad + (\Gamma_{2}' X_{t-278}^{t-290} + \Lambda_{2}' \epsilon_{t-278}^{t-290}) \notag \\
& \quad + (\Gamma_{3}' X_{t-360}^{t-372} + \Lambda_{3}' \epsilon_{t-360}^{t-372}) + \epsilon_t  \notag
\end{align}
More details of this model will be discussed in Section~\ref{subsec_face}.

For estimating coefficients $\Omega = \{\Phi,\Psi,\sigma,\Gamma_{i},\Lambda_{i} \text{ for } 1 \leq i \leq r\}$, maximum likelihood estimation (MLE) is utilized. In our package, the default optimization algorithm is BFGS~\cite{10.1093/imamat/6.1.76,10.1093/comjnl/13.3.317,10.2307/2004873,10.2307/2004840}. The formula of log-likelihood function and its gradient is given below.
\begin{flalign}
\text{Objective:} && \text{Minimize } l &= \sum_{t=1}^N \epsilon_t^2& \notag \\
&& \epsilon_t = X_t - \Phi' X_{t-1}^{t-p} &- \Psi' \epsilon_{t-1}^{t-q} - \sum_{i=1}^{r} (\Gamma_{i}' X_{\textbf{S}_i} + \Lambda_{i}' \epsilon_{\textbf{S}_i})& \notag \\
\text{Gradient:}&&\frac{\partial l}{\partial \Phi} = 2\sum_{t=1}^N \epsilon_t \frac{\partial \epsilon_t}{\partial \Phi} &\text{ , }
\frac{\partial l}{\partial \Psi} = 2\sum_{t=1}^N \epsilon_t \frac{\partial \epsilon_t}{\partial \Psi}& \notag \\
&&\frac{\partial l}{\partial \Gamma} = 2\sum_{t=1}^N \epsilon_t \frac{\partial \epsilon_t}{\partial \Gamma} &\text{ , }
\frac{\partial l}{\partial \Lambda} = 2\sum_{t=1}^N \epsilon_t \frac{\partial \epsilon_t}{\partial \Lambda}& \notag 
\\
\text{Where}&&\frac{\partial \epsilon_t}{\partial \Phi} = - X_{t-1}^{t-p} -& \Psi' \frac{\partial \epsilon_{t-1}^{t-q}}{\partial \Phi} - \sum_{i=1}^{r} \Lambda_i' \frac{\partial \epsilon_{\textbf{S}_i}}{\partial \Phi}& \notag \\
&&\frac{\partial \epsilon_t}{\partial \Psi} = - \epsilon_{t-1}^{t-q} -& \Psi' \frac{\partial \epsilon_{t-1}^{t-q}}{\partial \Psi} - \sum_{i=1}^{r} \Lambda_i' \frac{\partial \epsilon_{\textbf{S}_i}}{\partial \Psi}& \notag \\
&&\frac{\partial \epsilon_t}{\partial \Gamma_i} = - X_{\textbf{S}_i} -& \Psi' \frac{\partial \epsilon_{t-1}^{t-q}}{\partial \Gamma_i} - \sum_{i=1}^{r} \Lambda_i' \frac{\partial \epsilon_{\textbf{S}_i}}{\partial \Gamma_i}& \notag \\
&&\frac{\partial \epsilon_t}{\partial \Lambda_i} = - \epsilon_{\textbf{S}_i} -& \Psi' \frac{\partial \epsilon_{t-1}^{t-q}}{\partial \Lambda_i} - \sum_{i=1}^{r} \Lambda_i' \frac{\partial \epsilon_{\textbf{S}_i}}{\partial \Lambda_i}& \nonumber
\end{flalign}

\subsection{Modeling Procedure}

Based on the generalized seasonal ARIMA model, for a given $r,\tau, \bar p, \bar q$, we introduce the following Multiple Seasonality (MS) modeling procedure.

\begin{enumerate}
	\item Detect seasonality candidates $\textbf{S}_i$'s by spectrum analysis of $X_t$. The time series will be decomposed into $(N/2)$'s trigonometric components by Discrete Fourier Transformation (DFT)~\cite{dft}. 
	\begin{equation}
	X_t = a_0 + \sum_{j=1}^{N/2} [a_j cos(2\pi tj/N) + b_j sin(2\pi tj/N)] \notag
	\end{equation}
	Each trigonometric component is defined on an unique frequency $2\pi j/N$ assigned with a signal strength $P(j/N)=a_j^2 + b_j^2$. $r+2$'s frequencies with the largest signal strength will be selected as candidates in $C$.
	\item At each parameter combination $\{p,q,C\}$, fit multiple seasonality model by MLE. Calculate the information criterion for each combination. Bridge criterion (BC)~\cite{DingBridge} is employed as the default information criterion in the package.
	Then the combination of parameters is selected by optimizing the information criterion. Based on optimal model, perform forecast by $\hat X_{N+j} = E[X_{N+j}|X_{1}^N]$.
\end{enumerate}

The MS procedure's performance depends on the pre-specification of $r,\tau, \bar p, \bar q$. In practice, $\tau$ should be considerably large to allow flexibility in fitting. The default value for $\tau$ in our developed package (and experiments) is 6. $\bar p, \bar q$ shouldn't be too large. 
In practice, allowing overly large $\bar p, \bar q$ will overweight the importance of short-term dynamics and neglect potential seasonal effects. From various experiments, we found that the number of seasonality components $r$ is the most influential parameter to the model, and it deserves sophisticated tuning.

\subsection{Discussion}
In the sequel, we elaborate on some other aspects of the above model and procedure.
\begin{itemize}
	\item Overfitting. Since our method has a linear ARMA structure and allows limited ranges of ARMA components, it is unlikely to have an overfitting issue.
	\item Detection \& forecasting? The idea of our work is to propose a new modeling procedure to perform prediction, which takes into account multiple seasonality. The detection of seasonality is part of the method. We select seasonality by comparing the information criterion. While many state-of-the-art statistical methods such as TBATS, the Prophet model relies on pre-specification of seasonality, our method detects and suggests appropriate seasonality over a range of potential seasonality orders. As a result, our method performs well for multiple seasonality data scenarios where the seasonality pattern is not visually straightforward.
	\item Relation to ARIMA. The ARIMA model is an extension of the ARMA model when the data series are taken differencing operations. A usual ARIMA model can be expanded to ARMA expression. The point of introducing the ARIMA structure in the first place is to maintain the stationarity assumption of the ARMA model. Since we do not assume stationarity in our method, we only discussed the relationship between our method to the ARMA model.
	\item Stationarity. Since the essence of seasonality is the recurrence of long-term signals, 
	a seasonal time series model is generally non-stationary, e.g. the model $X_t = X_{t-12} + \epsilon_t$. 
	\item Theory derivation. Because we do not assume stationarity in the time series, it is typically difficult to analyze the model's theoretical property unless the time series is assumed to be decomposed into a periodic part and stationary part or by assuming local stationarity~\cite{wavelet, DingEvolution}.
	\item Identifiability. As $\tau,\bar p, \bar q$ increase, the identifiability of the model will be questionable. For example, the true model $X_t = X_{t-12} + \epsilon_t$ 
	may be also written as $X_t = X_{t-24} + \epsilon_{t-12} +\epsilon_t$. In general, we require that $\tau > \max\{\bar{p}, \bar{q}\}$.
	\item Information criterion. As stated in the last subsection, BC is taken as the default information criterion for the model selection, because of its theoretical advantages comparing to standard AIC or BIC ~\cite{DingBridge}. Nevertheless, we repeat our numerical and empirical studies with AIC and BIC as information criterion as well. They show similar results.
	\item ``Detrended'' Data. Currently, MS modeling is only suitable for detrended data. As future work, a component accommodating trend effects will be included in the formulation.
	\item Ultra-long term prediction. As a consequence of recurrent linear structure, the $n$-step ahead prediction will either converge or vibrate drastically when $n$ is very large. A suitable correction technique may be used for ultra long term prediction~\cite{effective}. From our experiments, this is not an issue if $n<100$ for $N=1000$.
	\item Computational complexity for long series. Since the MLE involves the computation of gradient recursively, the computational costs cumulate as the length of series increases. The computational complexity follows the order of $O(N)$, where $N$ is the sample size. The track of the computational time of MS procedure in single seasonality and double seasonality with trigonometric components in Section~\ref{simulation} is shown below.
	
	\begin{figure}[ht]
		\vskip 0.1in
		\begin{center}
			\centerline{\includegraphics[width=\columnwidth]{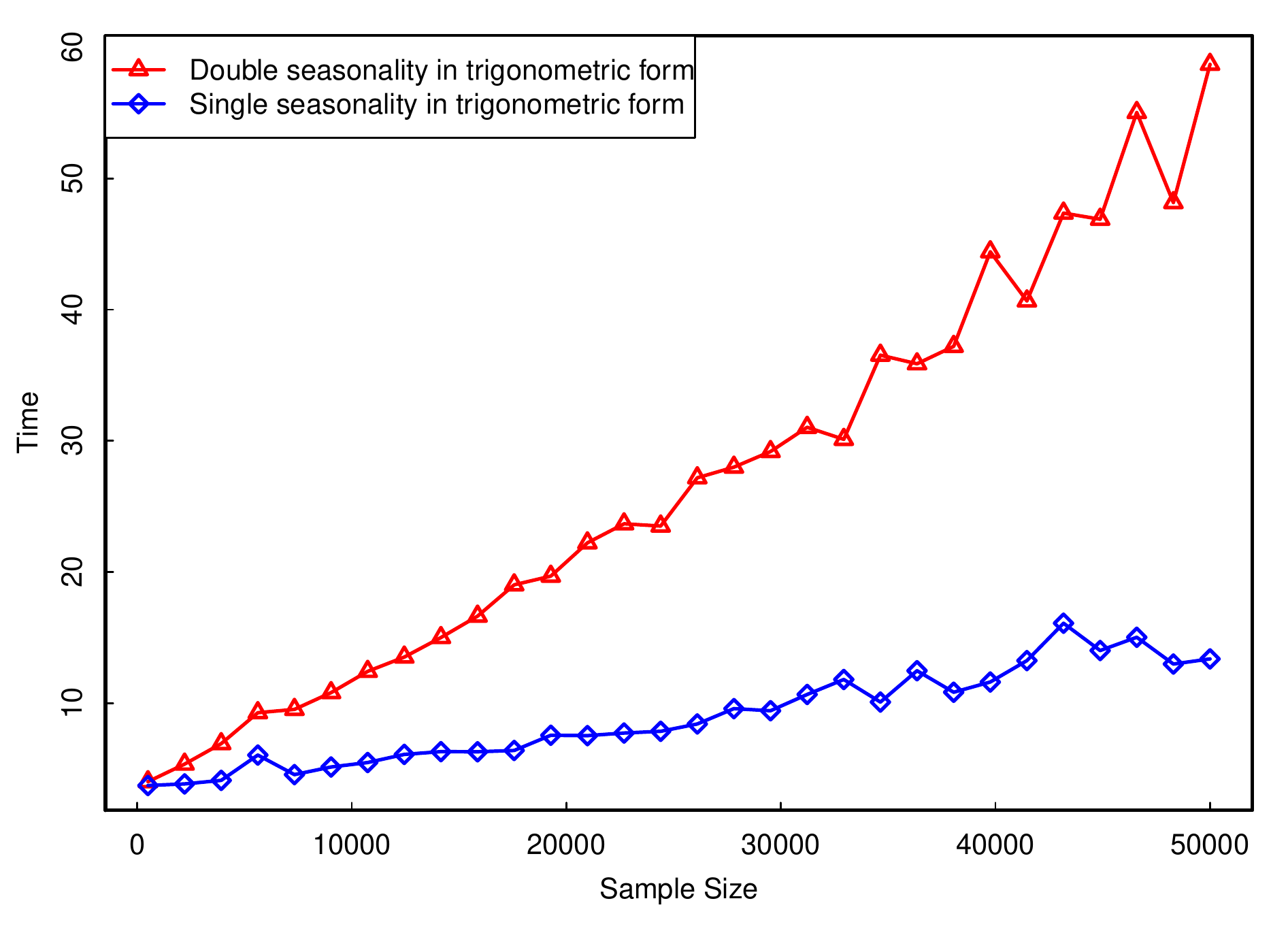}}
			\vskip -0.1in
			\caption{Computational time in two different data setups. As the figure has shown, the computational time goes up as the sample size increases. Both of the curves exhibit a linear relation between computational time and sample size.}
		\end{center}
		\vskip -0.1in
	\end{figure}
	
\end{itemize}


\section{Experimental Studies}
In this section, we evaluate the forecasting capabilities of the proposed method through various experimental studies.

\subsection{Numerical Examples} \label{simulation}

In the following experiment studies, we will compare our MS modeling procedure to benchmark methods, including the vanilla ARMA model selected by BC, the Facebook Prophet model, the TBATS model (both with correct pre-determined seasonality orders) and a two-layer LSTM with 50 neurons on each layer. The LSTM model is a simple LSTM model recommended in the Pytorch tutorial for sin-wave time sequence prediction. We use this LSTM model because we try to compare our method with benchmark methods with similar complexity and computational cost. We train it individually in each repetition on each series by Adam optimizer, with learning rate 0.01 and 600 iterations. 

The simulation includes four different data setups. In each setup, 30 parallel univariate sequences are generated, and the models are evaluated 50 times on each sequence. Specifically, in the $i$-th evaluation where $1\leq i \leq 50$, models will be trained on the first $650 + 5*(i-1)$ observations from the sequence, and predict the following 100. The mean value and standard error of CMSE for n-step ahead prediction is reported.

In the first data setup, we consider single seasonality in trigonometric form. The reason to start with single seasonality is that most literature of forecasting time series with seasonality focus on single seasonality. We need to ensure that our MS procedure has satisfactory performance in a simple situation. The data is generated as an aggregation of a trigonometric component and a short-term random effect, which is the standard way to simulate time series with seasonality.

The first column refers to different prediction length $n$. For each value of $n$, the mean value of cumulative mean squared error is recorded in the first row, and the standard error is recorded in parenthesis. Mean value with distance to smallest one in a row less than the summation of their standard errors, will be highlighted.

\begin{align}
X_t &= 10 sin(\frac{2\pi}{50}t) + Z_t \notag \\
Z_t &= 0.8 Z_{t-1} - 0.3 Z_{t-2} + 0.5 \epsilon_{t-1} + \epsilon_t \notag\\
\epsilon_t &\sim^{i.i.d} N(0,\sigma^2) \text{ where } \sigma=2
\end{align}
\begin{table}[h]
	\caption{Single seasonality in trigonometric form.}
	\begin{center}
		\begin{small}
			\begin{sc}
				\begin{tabular}{l|ccccr}
					\toprule
					& ARMA & MS & Prophet & LSTM & TBATS\\ 
					\midrule 
					$n=$1 & 5.85 & \textbf{4.61} & 15.45 & 6.55 & \textbf{4.64}\\ 
					& (0.38) & (0.29) & (0.95) & (0.40) & (0.29)\\ 
					$n=$5 & 26.01 & \textbf{12.16} & 14.67 & 24.33 & \textbf{12.01} \\ 
					& (1.23) & (0.53) & (0.59) & (1.22) & (0.53)\\ 
					$n=$15 & 49.14 & \textbf{14.28} & \textbf{14.64} & 32.74 & \textbf{13.91}\\ 
					& (1.38) & (0.38) & (0.37) & (1.27) & (0.37)\\ 
					$n=$50 & 61.91 & 15.21 & \textbf{14.78} & 40.97 & \textbf{14.69}\\ 
					& (0.70) & (0.22) & (0.23) & (1.34) & (0.22)\\ 
					$n=$100 & 64.25 & 15.64 & \textbf{15.09} & 50.26 & \textbf{15.09}\\ 
					& (0.49) & (0.17) & (0.17) & (1.60) & (0.17)\\ 
					\bottomrule
				\end{tabular} 
			\end{sc}
		\end{small}
	\end{center}
\end{table}

The results show that MS has the best performance for 1-step ahead prediction. It also provides comparable results to TBATS in long-term prediction, where $n=5,15$. MS also outperforms Prophet when $n=1,5,15$.

In the second data setup, we consider double seasonality in trigonometric form. Since double seasonality is the simplest situation in multiple seasonality regimes, we will emphasize simulation in a double seasonality situation. The data is generated as an aggregation of two trigonometric components with different frequencies and a short-term random effect.

\begin{align}
X_t &= 10 sin(\frac{2\pi}{50}t) + 5 sin(\frac{2\pi}{15}t)+ Z_t \notag \\
Z_t &= 0.8 Z_{t-1} - 0.3 Z_{t-2} + 0.5 \epsilon_{t-1} + \epsilon_t \notag\\
\epsilon_t &\sim^{i.i.d} N(0,\sigma^2) \text{ where } \sigma=2
\end{align}
\begin{table}[h]
	\caption{Double Seasonality in trigonometric form.}
	\begin{center}
		\begin{small}
			\begin{sc}
				\begin{tabular}{l|ccccr}
					\toprule
					& ARMA & MS & Prophet & LSTM & TBATS \\ 
					\midrule
					$n=$1 & 5.37 & \textbf{4.29} & 52.38 & 8.16 & \textbf{3.91}\\ 
					& (0.33) & (0.28) & (2.89) & (0.52) & (0.25)\\ 
					$n=$5 & 33.78 & 14.69 & 50.28 & 37.40 & \textbf{11.70}\\ 
					& (1.65) & (0.70) & (1.52) & (2.05) & (0.48)\\ 
					$n=$15 & 61.41 & 16.89 & 50.28 & 58.69 & \textbf{13.35}\\ 
					& (1.74) & (0.47) & (0.88) & (2.41) & (0.32)\\ 
					$n=$50 & 72.74 & 18.43 & 50.09 & 74.75 & \textbf{13.91}\\ 
					& (0.73) & (0.34) & (0.42) & (2.10) & (0.18)\\ 
					$n=$100 & 74.34 & 19.46 & 49.52 & 90.63 & \textbf{13.73}\\ 
					& (0.50) & (0.33) & (0.31) & (2.18) & (0.12)\\ 
					\bottomrule
				\end{tabular}
			\end{sc}
		\end{small}
	\end{center}
\end{table}

The results show that TBATS is better than MS for most of the prediction length. The results are conceivable because TBATS utilize trigonometric components to accommodate seasonality. Naturally, it is suitable for modeling seasonality generated in trigonometric form. Furthermore, seasonality orders of TBATS are correctly pre-determined. However, the Prophet model performs poorly due to the disturbance from short term random effect, although it also utilizes trigonometric components for seasonality. MS outperforms the Prophet model at all prediction lengths.

Therefore, to further explore the methods' capability of modeling seasonality in a complex form, we consider non-trigonometric double seasonality in the third data setup. The seasonality component comes from the repetition of exogenous white-noise series instead of trigonometric functions. Then the observed data is generated as an aggregation of two non-trigonometric components with different frequencies and a short-term random effect.

\begin{align}
X_t &= A_t + B_t + Z_t \notag \\
Z_t &= 0.8 Z_{t-1} - 0.3 Z_{t-2} + 0.5 \epsilon_{t-1} + \epsilon_t \notag\\
\epsilon_t &\sim^{i.i.d} N(0,\sigma^2) \text{ where } \sigma=2 \notag 
\end{align}
\begin{align}
A_t = \begin{cases}
a_t \sim^{i.i.d} N(0,1) &\text{ for } 0 < t < 100\\
A_{t-50} &\text{ for } 100\leq t
\end{cases} \notag \\
B_t = \begin{cases}
b_t \sim^{i.i.d} N(0,1) &\text{ for } 0 < t < 100\\
B_{t-15} &\text{ for } 100\leq t
\end{cases} 
\end{align}
\begin{table}[h]
	\caption{Double Seasonality in non-trigonometric form.}
	\begin{center}
		\begin{small}
			\begin{sc}
				\begin{tabular}{l|ccccr}
					\toprule
					& ARMA & MS & Prophet & LSTM & TBATS \\ 
					\midrule
					$n=$1 & 76.30 & \textbf{31.80} & 68.04 & 149.76 & 45.32\\ 
					& (4.09) & (2.50) & (3.98) & (9.98) & (2.73)\\ 
					$n=$5 & 80.05 & \textbf{44.01} & 75.30 & 151.37 & \textbf{46.60}\\ 
					& (2.36) & (1.52) & (2.13) & (4.75) & (1.35)\\ 
					$n=$15 & 80.88 & \textbf{47.58} & 75.72 & 150.99 & \textbf{48.72}\\ 
					& (1.35) & (1.05) & (1.23) & (2.96) & (0.82)\\ 
					$n=$50 & 82.06 & 51.98 & 76.39 & 156.00 & \textbf{49.70}\\ 
					& (0.51) & (0.83) & (0.45) & (1.77) & (0.42)\\ 
					$n=$100 & 83.48 & 54.55 & 77.20 & 158.35 & \textbf{50.09}\\ 
					& (0.36) & (0.81) & (0.34) & (1.46) & (0.36)\\ 
					\bottomrule
				\end{tabular}
			\end{sc}
		\end{small}
	\end{center}
\end{table}

The results show that MS have superiority when $n=1,5,15$, and TBATS model has best performance when $n=50,100$. MS is better than the Prophet at all prediction length. Overall, comparing to modeling trigonometric seasonality, MS has more potential for modeling non-trigonometric seasonality.

In the fourth data setup, we add up both trigonometric and non-trigonometric components for double seasonality. The complexity of seasonality structure increased one more time.

\begin{align}
X_t &= 10 sin(\frac{2\pi}{50}t) + 5 sin(\frac{2\pi}{15}t)+ A_t + B_t + Z_t \notag \\
Z_t &= 0.8 Z_{t-1} - 0.3 Z_{t-2} + 0.5 \epsilon_{t-1} + \epsilon_t \notag\\
\epsilon_t &\sim^{i.i.d} N(0,\sigma^2) \text{ where } \sigma=2 \notag 
\end{align}
\begin{align}
A_t = \begin{cases}
a_t \sim^{i.i.d} N(0,1) &\text{ for } 0 < t < 100\\
A_{t-50} &\text{ for } 100\leq t
\end{cases} \notag \\
B_t = \begin{cases}
b_t \sim^{i.i.d} N(0,1) &\text{ for } 0 < t < 100\\
B_{t-15} &\text{ for } 100\leq t
\end{cases} 
\end{align}
\begin{table}[h]
	\caption{Double Seasonality with both trigonometric and non-trigonometric components.}
	\begin{center}
		\begin{small}
			\begin{sc}
				\begin{tabular}{l|ccccr}
					\toprule
					& ARMA & MS & Prophet & LSTM & TBATS \\ 
					\midrule
					$n=$1 & 102.60 & \textbf{17.94} & 108.33 & 146.58 & 43.47\\ 
					& (6.80) & (1.24) & (7.07) & (9.38) & (2.67)\\ 
					$n=$5 & 125.52 & \textbf{34.50} & 112.28 & 174.06 & 47.02\\ 
					& (3.98) & (1.39) & (3.14) & (5.54) & (1.46)\\ 
					$n=$15 & 141.14 & \textbf{39.32} & 112.09 & 186.63 & 49.49\\ 
					& (2.54) & (0.94) & (1.75) & (4.12) & (0.88)\\ 
					$n=$50 & 147.15 & \textbf{46.09} & 110.09 & 206.93 & 50.29\\ 
					& (1.19) & (0.79) & (0.90) & (2.92) & (0.51)\\ 
					$n=$100 & 147.89 & \textbf{48.62} & 110.09 & 223.90 & 50.37\\ 
					& (0.88) & (0.82) & (0.73) & (2.79) & (0.46)\\ 
					\bottomrule 
				\end{tabular}
			\end{sc}
		\end{small}
	\end{center}
\end{table}
The results show that MS outperforms TBATS and Prophet at all prediction lengths, especially for short-term prediction. MS is suitable for modeling multiple seasonality in complex form.

Generally speaking, in the simulation study above, MS modeling procedure shows comparable performance to the benchmark method such as TBATS and outperforms method like Prophet in most cases. MS also shows its suitability for modeling multiple seasonality in non-trigonometric form, which might be potentially useful in real data application, since the seasonality structure is usually more complicated than trigonometric functions in real data. Therefore, in the next subsection, we will evaluate MS procedure and other methods on four real-data case study, to further show its capability of modeling multiple seasonality.

\subsection{Empirical Studies}

\subsubsection{PJM Load Forecasting}
In many countries worldwide, electricity is now traded under market rules using spot and derivative contracts~\cite{energybook}. At the corporate level, electricity load and price forecasts have become a fundamental input to energy companies’ decision making mechanisms. The costs of over- or undercontracting and then selling or buying power in the balancing market are typically so high that they can lead to huge financial losses and bankruptcy in the extreme case.~\cite{forecastbook} The risk goes up when pass on to consumers. A failure in forecasting electricity utility can eventually lead to disastrous power outage. (e.g. Manhattan blackout in July 2019)

This case study focuses on forecasting electricity load for PJM, a regional transmission organization (RTO) that coordinates the movement of wholesale electricity in all or parts of 13 states and District of Columbia in US. The data is the hourly load of electricity across three regional company under PJM from 9/30/2018 to 10/1/2019. (PEPCO : Potomac Electric Power Company; PE : Pennsylvania Electric Company; COMED : ComEd) Forecasting will be performed weekly for the last five weeks on rolling basis. In this case, daily (period as 24) and weekly (period as 24*7) seasonality terms are specified in Prophet and TBATS model. Two seasonality components are allowed in MS procedure ($r=2$).

\begin{figure}[h]
	\begin{center}
		\centerline{\includegraphics[width=\columnwidth]{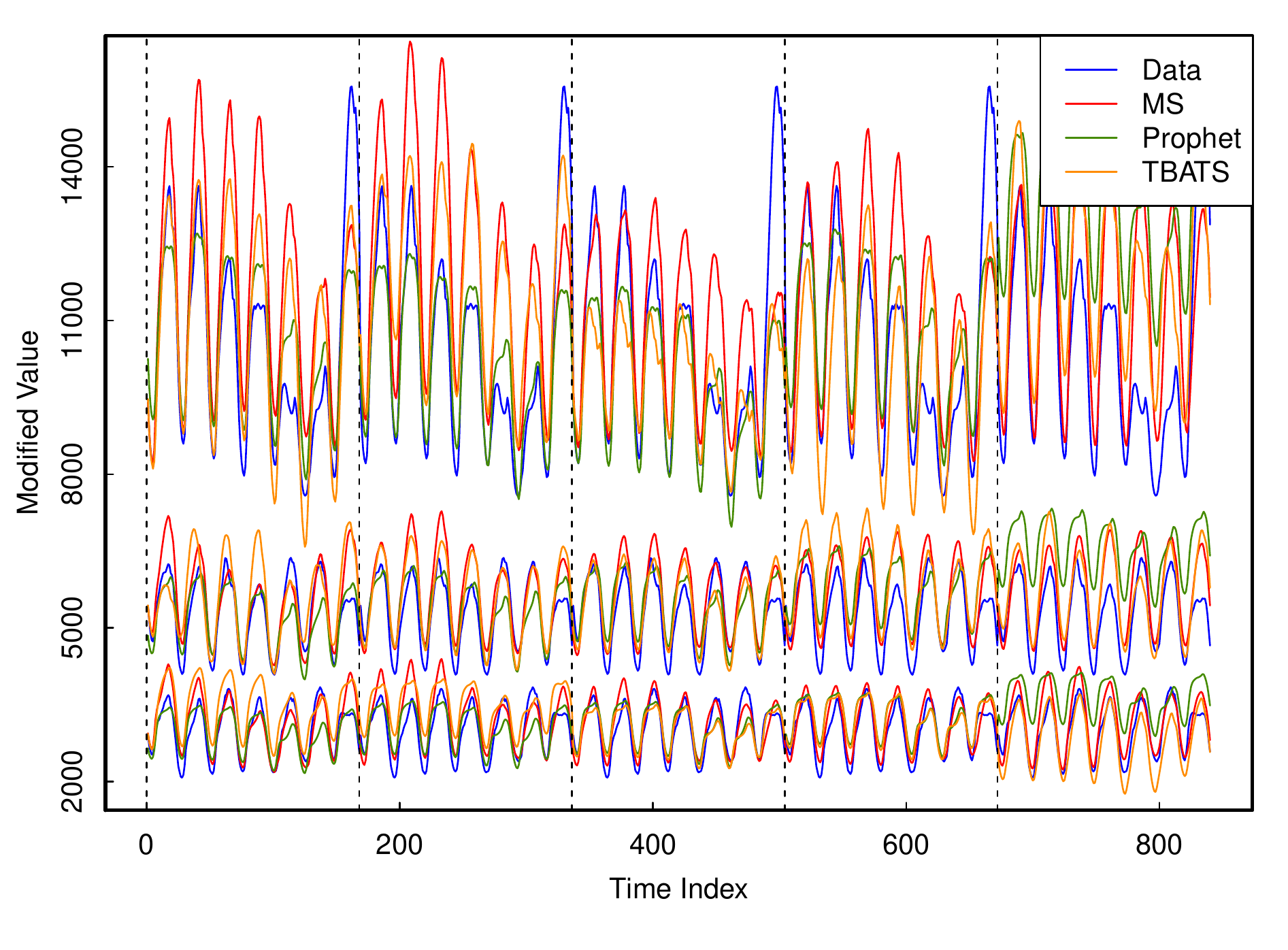}}
		\caption{Predicted curves of PJM load data. The values shown in the figure are modified to separate the curves. From the top to the bottom, the three bunches of curves correspond to PEPCO, PE and COMED accordingly. MS, Prophet and TBATS are evaluated 5 times for forecasting the following 24*7 steps. Each interval bounded by black dashed lines corresponds to an evaluation. Specifically, in each interval, models are trained by data before the left line, and forecast till the right line.}
	\end{center}
\end{figure}

\begin{table}[h]
	\caption{Mean and SE of standardized CMSE.}
	\begin{center}
		\begin{small}
			\begin{sc}
				\begin{tabular}{l|ccccr}
					\toprule
					& $n=$1 & $n=$12 & $n=$24 & $n=$72 & $n=$168 \\ 
					\midrule
					MS& 0.30 & \textbf{0.16} & \textbf{0.30} & \textbf{0.51} & \textbf{0.59} \\
					& (0.12) & (0.04) & (0.12) & (0.14) & (0.13) \\ 
					Prophet& 0.47 & 0.50 & 0.55 & 0.80 & 0.85 \\
					& (0.31) & (0.38) & (0.37) & (0.52) & (0.47) \\
					TBATS & 0.23 & 0.23 & 0.38 & 0.60 & 0.66 \\
					& (0.15) & (0.12) & (0.13) & (0.19) & (0.13) \\
					\bottomrule
				\end{tabular}
			\end{sc}
		\end{small}
	\end{center}
\end{table}

The MS procedure indicates the seasonality terms at periods 21-27, 163-171, matching daily and weekly seasonality. It agrees with common knowledge of electricity load pattern. As it shows from the figure and table, MS procedure outperforms Prophet and TBATS model for most prediction lengths.

\subsubsection{Total Sun Spot Number}\label{sub_sunspot}

This is a case study of monthly total sun spot number from 1749 to 2019. Existence of seasonality pattern in this univariate time series is well known. The seasonality period range from 9 to 14 years in history, and has averaged length as 11 years for last decades. MS, Prophet and TBATS model are evaluated 5 times, forecasting the following 150 steps. The seasonality period for Prophet and TBATS model is specified as 132 (11 years). Two seasonality components are allowed in MS ($r=2$).

\begin{figure}[h]
	\begin{center}
		\centerline{\includegraphics[width=\columnwidth]{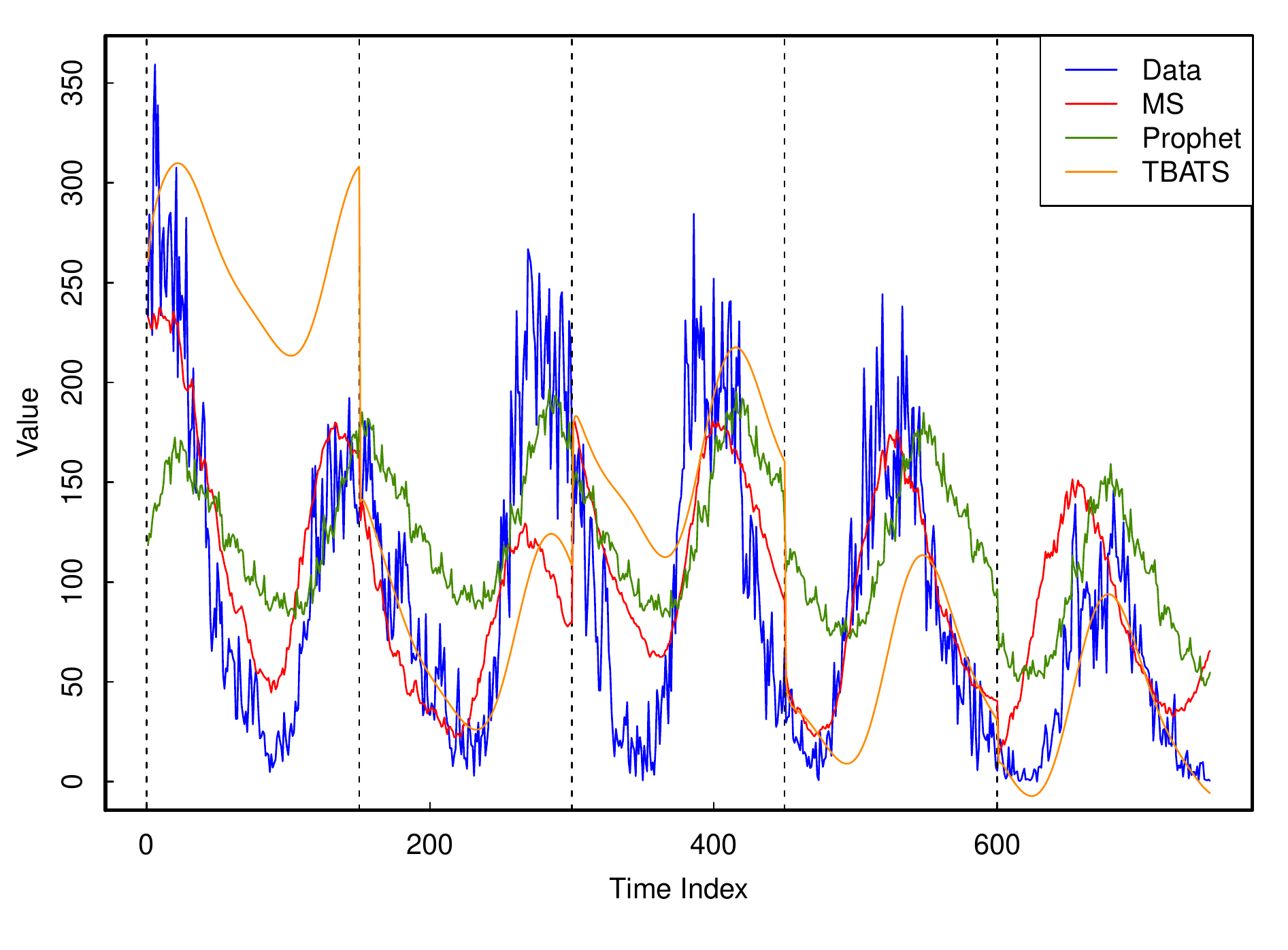}}
		\caption{Predicted curves of total sun spot number. MS, Prophet and TBATS are evaluated 5 times for forecasting the following 150 steps. Each interval bounded by black dashed lines corresponds to an evaluation.}
	\end{center}
\end{figure}

\begin{table}[h]
	\caption{Mean and SE of standardized CMSE.}
	\begin{center}
		\begin{small}
			\begin{sc}
				\begin{tabular}{l|ccccr}
					\toprule
					& $n=$1 & $n=$5 & $n=$25 & $n=$50 & $n=$100 \\ 
					\midrule
					MS& \textbf{8.26} &9.70 & \textbf{14.87} & \textbf{23.08} & \textbf{19.52} \\
					& (6.79) & (5.0) & (4.64) & (7.61) & (5.04) \\ 
					Prophet& 47.71 & 62.40 & 59.32 & 45.67 & 41.62 \\
					& (22.52) & (38.87) & (27.25) & (12.41) & (7.97) \\
					TBATS & 9.96 & 6.57 & 15.0 & 46.25 & 78.79 \\
					& (6.67) & (2.81) & (8.01) & (23.55) & (46.16) \\
					\bottomrule
					\multicolumn{6}{l}{The values are reported at $10^2$ scale.}
				\end{tabular}
			\end{sc}
		\end{small}
	\end{center}
\end{table}

MS in this example suggests a range of seasonality period at 8-12 years instead of picking a single value. As it also shows from the figure and table, MS outperforms other two for most prediction lengths. This case reveals suitability of MS for modeling time series with unclear seasonality pattern. By allowing a range instead of fixing seasonality period, the model is able to adjust periodic pattern along the time series, creating more flexibility.

\subsubsection{Air Quality (CO Level)}
The data \cite{airquality} contains the responses of a gas multi-sensor device deployed on the field in an Italian city. Hourly responses averages are recorded along with gas concentrations references from a certified analyzer. In this case study, we focus on forecasting CO level. MS, Prophet and TBATS are evaluated 10 times for forecasting the following 50 steps. Seasonality period as 24 is specified for Prophet and TBATS. Two seasonality components are allowed in MS ($r=2$).

\begin{figure}[h]
	\begin{center}
		\centerline{\includegraphics[width=\columnwidth]{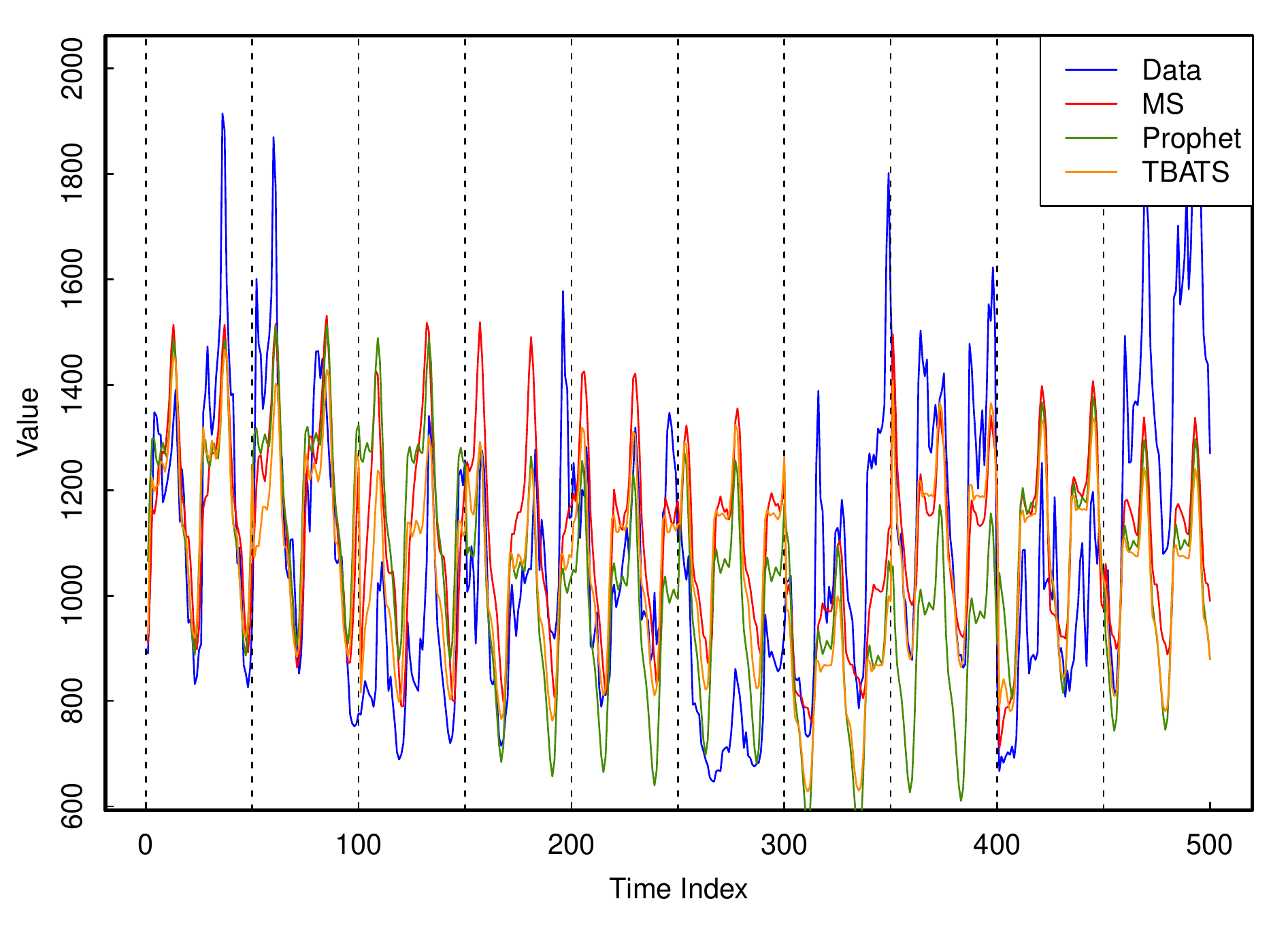}}
		\vskip -0.1in
		\caption{Predicted curves of air quality (CO level). MS, Prophet and TBATS are evaluated 10 times for forecasting the following 50 steps. Each interval bounded by black dashed lines corresponds to an evaluation.}
	\end{center}
\end{figure}

\begin{table}[h]
	\caption{Mean and SE of standardized CMSE.}
	\begin{center}
		\begin{small}
			\begin{sc}
				\begin{tabular}{l|ccccr}
					\toprule
					& $n=$1 & $n=$5 & $n=$12 & $n=$24 & $n=$48 \\ 
					\midrule
					MS & \textbf{10.04} & 27.20 & 35.87 & 42.18 & \textbf{51.94} \\
					& (5.75) & (8.32) & (11.50) & (9.63) & (12.87)\\
					Prophet & 60.05 & 52.18 & 47.47 & 52.75 & 65.63 \\
					& (25.34) & (20.86) & (18.99) & (14.31) & (15.42)\\
					TBATS & 11.12 & 24.63 & 26.96 & 39.56 & 52.76 \\
					& (5.51) & (10.43) & (9.82) & (10.34) & (16.69) \\
					\bottomrule
					\multicolumn{6}{l}{The values are reported at $10^3$ scale.}
				\end{tabular} 
			\end{sc}
		\end{small}
	\end{center}
\end{table}

The MS procedure indicates seasonality periods at 18-30, 37-45. The former one corresponds to daily seasonality period, since the median of 18-30 is 24. The second one can be referring to the effect from two days ago. For this single seasonality data situation, MS provide satisfactory results, and outperforms other two for $n=1,48$.

\subsubsection{Facebook Events Data}\label{subsec_face}
The data comes from the original paper of Facebook Prophet model~\cite{prophet}. It shows daily data for the number of events created on Facebook during the dates from 12/10/2007 to 01/20/2016. Weekly and yearly seasonality are strongly suspected. In this case study, MS,Prophet and TBATS are evaluated 7 times for forecasting the following 200 steps. Weekly and yearly seasonality periods are specified for Prophet and TBATS model, and three seasonality components are allowed in MS ($r=3$).

\begin{figure}[h]
	\begin{center}
		\centerline{\includegraphics[width=\columnwidth]{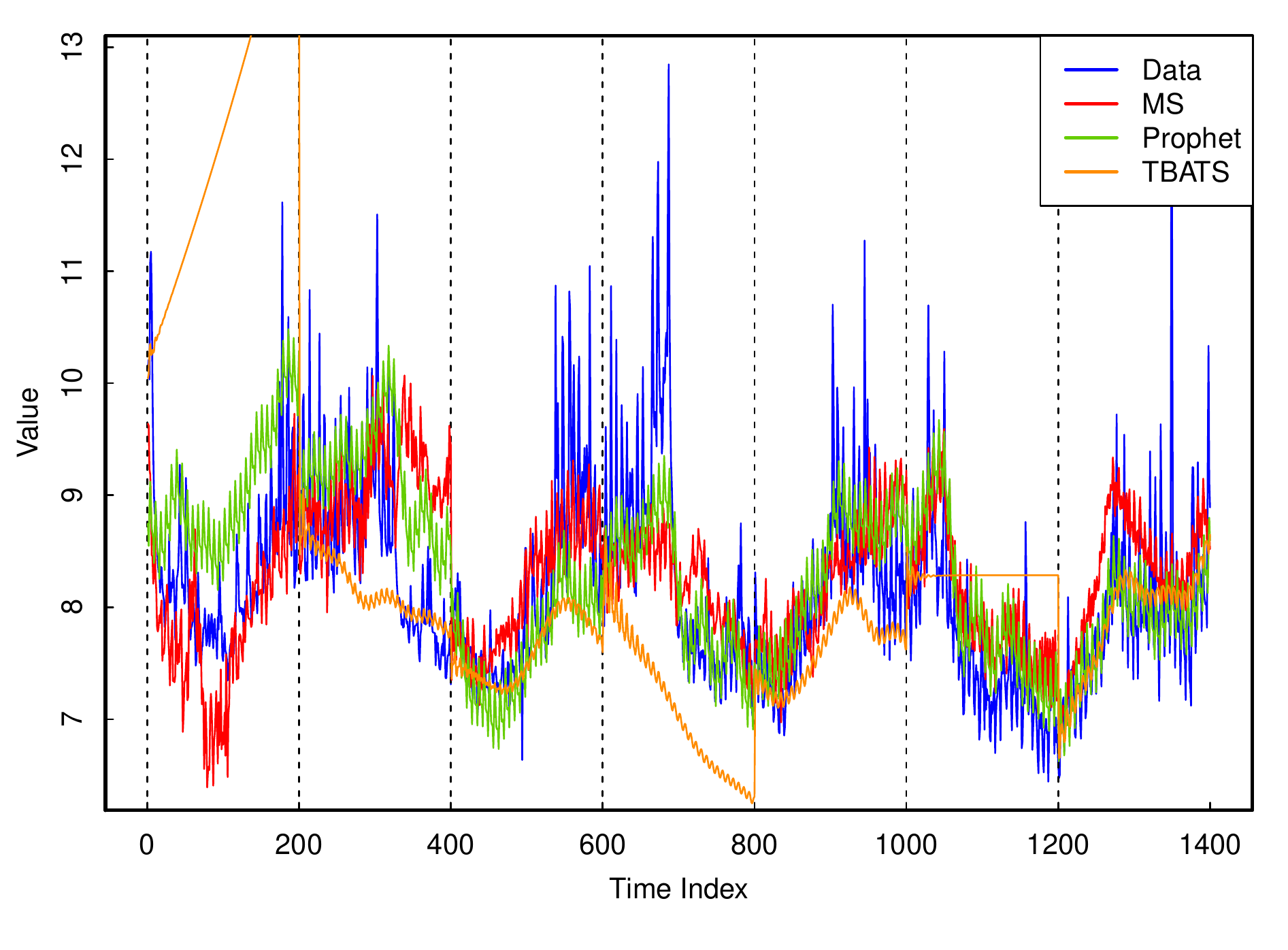}}
		\vskip -0.1in
		\caption{Predicted curves of Facebook events data. MS, Prophet and TBATS are evaluated 7 times for forecasting the following 200 steps. Each interval bounded by black dashed lines corresponds to an evaluation. }
	\end{center}
\end{figure}

\begin{table}[h]
	\caption{Mean and SE of standardized CMSE.}
	\begin{center}
		\begin{small}
			\begin{sc}
				\begin{tabular}{l|ccccr}
					\toprule
					& $n=$1 & $n=$10 & $n=$50 & $n=$100 & $n=$200 \\ 
					\midrule
					MS & \textbf{0.11} & 0.43 & 0.34 & 0.48 & 0.57 \\ 
					& (0.05) & (0.26) & (0.08) & (0.15) & (0.10)\\
					Prophet & 0.45 & 0.46 & 0.31 & 0.37 & 0.51\\ 
					& (0.27) & (0.29) & (0.10) & (0.11) & (0.09)\\
					TBATS & 0.13 & 0.24 & 1.06 & 2.18 & 3.31 \\
					& (0.10) & (0.08) & (0.71) & (1.39) & (2.41) \\
					\bottomrule
				\end{tabular}
			\end{sc}
		\end{small}
	\end{center}
\end{table}

The MS procedure indicates seasonality period range at 116-128, 278-290, 360-372, where the last one refer to yearly seasonality. MS provides comparable results to Prophet in terms of long term prediction, and has better performance for short term prediction. TBATS performs poorly for long-term prediction.

\section{Conclusion}
In this paper, we have introduced MS modeling procedure, a simple yet powerful forecasting procedure that combines seasonality detection, estimation, model selection and forecasting for time series with multiple seasonality. The procedure doesn't require any pre-determined seasonality periods, and has a limited number of parameters to specify. We reveal its competence of forecasting by experimental and empirical studies. In most cases, MS outperforms the state-of-the-art such as the Facebook Prophet model. 

\balance

\bibliographystyle{IEEEtran}
\bibliography{bigdata}

\end{document}